\documentclass[letterpaper, 10 pt, journal, twoside]{IEEEtran}
\ifCLASSINFOpdf
\else
\fi
\usepackage[dvipdfmx]{graphicx}
\usepackage[dvipdfmx]{color}
\usepackage{url}
\usepackage{standalone}
\usepackage{times}
\usepackage{float}
\usepackage{acronym}
\usepackage{bm}
\usepackage{amsmath,amssymb}
\usepackage{etoolbox}
\usepackage{algorithm}
\usepackage{algorithmic}
\usepackage{subfigure}

\newcommand{\pr}[1]{\subsubsection{#1}}

\DeclareMathOperator{\tr}{tr}

\DeclareMathOperator*{\argmin}{arg\,min}

\acrodef{MDP}{Markov Decision Process}
\acrodef{POMDP}{Partially Observable Markov Decision Process}
\acrodef{WO}{Wheel Odometry}
\acrodef{VO}{Visual Odometry}

\begin{document}
\title{Where to Map? Iterative Rover-Copter Path Planning for Mars Exploration}

%\author{Michael~Shell,~\IEEEmembership{Member,~IEEE,}
%       John~Doe,~\IEEEmembership{Fellow,~OSA,}
%        and~Jane~Doe,~\IEEEmembership{Life~Fellow,~IEEE}% <-this % stops a space
%\thanks{M. Shell was with the Department
%of Electrical and Computer Engineering, Georgia Institute of Technology, Atlanta,

%\author{Takahiro Sasaki\inst{1} \and Kyohei Otsu \and Rohan Thakker\inst{2} \and Sofie Haesaert\inst{3} \and Ali-akbar Agha-mohammadi\inst{2}}

\author{Takahiro Sasaki$^{1}$, Kyohei Otsu$^{2}$, Rohan Thakker$^{2}$, Sofie Haesaert$^{3}$, and Ali-akbar Agha-mohammadi$^{2}$
% <-this % stops a space
\thanks{Manuscript received: September 10, 2019; Revised: December 6, 2019; Accepted: December 28, 2019.}%Use only for final RAL version
\thanks{This paper was recommended for publication by Editor Jonathan Roberts upon evaluation of the Associate Editor and Reviewers' comments.}
\thanks{The research was carried out at the Jet Propulsion Laboratory and the California Institute of Technology, under a contract with the National Aeronautics and Space Administration and funded through the President's and Director's Fund Program. U.S. Government sponsorship acknowledged.}% <-this % stops a space
\thanks{$^{1}$T. Sasaki is with Japan Aerospace Exploration Agency, Tsukuba, Japan,
({\small e-mail: sasaki.takahiro@jaxa.jp})}
\thanks{$^{2}$K. Otsu, R. Thakker, and A. Agha-Mohammadi are with Jet Propulsion Laboratory, California Institute of Technology, Pasadena, CA, ({\small e-mail: kyohei.otsu@jpl.nasa.gov; rohan.a.thakker@jpl.nasa.gov; aliakbar.aghamohammadi@jpl.nasa.gov})}
\thanks{$^{3}$S. Haesaert is with Eindhoven University of Technology, Eindhoven, The Netherlands, ({\small e-mail: s.haesaert@tue.nl})}
\thanks{Digital Object Identifier (DOI): see top of this page.}}


%\markboth{IEEE ROBOTICS AND AUTOMATION LETTERS,~Vol.~XX, No.~X, XXXXX~2020}{SASAKI \MakeLowercase{\textit{et al.}}: Where to Map? Iterative Rover-Copter Path Planning for Mars Exploration}
\markboth{IEEE Robotics and Automation Letters. Preprint Version. Accepted December, 2019}
{SASAKI \MakeLowercase{\textit{et al.}}: Where to Map? Iterative Rover-Copter Path Planning for Mars Exploration} 

\maketitle

\begin{abstract}
In addition to conventional ground rovers, the Mars 2020 mission will send a helicopter to Mars. 
The copter's high-resolution data helps the rover to identify small hazards such as steps and pointy rocks, as well as providing rich textual information useful to predict perception performance. In this paper, we consider a three-agent system composed of a Mars rover, copter, and orbiter. The objective is to provide good localization to the rover by selecting an optimal path that minimizes the localization uncertainty accumulation during the rover's traverse. 
To achieve this goal, we quantify the {\it localizability} as a goodness measure associated with the map, and conduct a joint-space search over rover's path and copter's perceptual actions given prior information from the orbiter. We jointly address {\it where to map} by the copter and {\it where to drive} by the rover using the proposed iterative copter-rover path planner.  We conducted numerical simulations using the map of Mars 2020 landing site to demonstrate the effectiveness of the proposed planner.
\end{abstract}


% \fi
\IEEEpeerreviewmaketitle

\section{Introduction}
\begin{figure}[t]
	\centering
	\includegraphics[width=0.75\columnwidth]{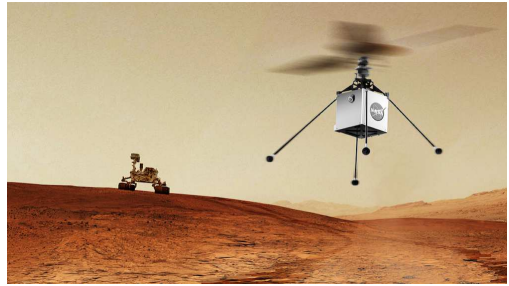}
	\vspace{-2mm}
	\caption{Rover/copter team for Mars exploration.}
	\vspace{-5mm}
	\label{fig:8_rc}
\end{figure}

\IEEEPARstart{T}{he} National Aeronautics and Space Administration (NASA) has sent several mobile rovers onto the surface of the red planet, to understand the Martian geology and the habitability of the environment. Recently, besides the conventional ground rovers, NASA decided to send a helicopter to the Martian sky, which will be traveling with a rover in the Mars 2020 mission \cite{Balaram2018} (Fig.~\ref{fig:8_rc}). Although the objective of this mission is limited to a technology demonstration (with less than 90 seconds flight time per day), future rover missions might be accompanied by copters, which work as low-flying scouts providing rich information about locations ahead of the rover. Such data provided by the copters could significantly increase the probability of mission success. 
Currently, the highest resolution achieved by a satellite is 25\,cm/px (HiRISE) \cite{HiRise}, which means small rocks remain undetected and are not taken into account by long-range strategic planners. These micro-structures could be critical hazards to the rover, such as sharp pointy rocks that significantly damaged the wheels in the Mars Science Laboratory (MSL) mission \cite{Arvidson2017}.
The copter's high-resolution data helps the rover to detect such small hazards in advance. In addition, it could provide richer information that further helps rover's path planning. For example, high-resolution top-view data will increase the confidence of localizability information with increased resolution and contextual information.
% POMDP
Autonomy will play an important role in future complex missions where multiple assets act independently\cite{Nilsson2018}. Localization, or more generally state estimation, is one of the key components in establishing autonomy. Due to the absence of global positioning systems, dead-reckoning methods such as \ac{WO} and \ac{VO} are the major source of localization on Mars. The rover's remote sensing capability is strongly impacted by localization performance since the accumulated errors in rover positions impose challenges in targeted observations after a few drives. 
Perception-aware planning is one of the effective methods to improve the performance of dead-reckoning-based localization \cite{SLAP, Inoue2016,Costante2017,IEEE:Kyon, SKKRAL}. It aims at improving the perception results by actively choosing future measurement targets. For example, the works in \cite{IEEE:Kyon,SKKRAL} improved the performance of \ac{VO} localization by actively choosing timing and camera direction to obtain an optimal image sequence using the predictive perception technique. This problem is typically approached by \ac{POMDP}, or belief-space planning \cite{POMDP1,POMDP2,bsp,FIRM, BVL}, where the planner chooses optimal actions under motion and sensing uncertainty.

% Our approach / Planning with two robots
In this paper, we consider a three-agent system composed of Mars rover, copter, and orbiter. The objective is to compute an optimal rover path that minimizes the localization uncertainty accumulation after a traverse. The key observation is that the accuracy of dead-reckoning localization depends on terrain types; for example, \ac{VO} accuracy is degraded in feature-poor terrain such as sand \cite{Maimone2007}. In other words, the localization performance can be boosted by selectively driving over ``good'' terrain to localize, which we call \textit{localizability} in the rest of the paper. Given the prior localizability map provided by remote satellite measurements, we conduct a joint-space search over rover's path and copter's perceptive actions while taking dynamic map updates from the copter into account. In other words, we jointly address \textit{where to map} by the copter and \textit{where to drive} by the rover in order to minimize the uncertainty accumulation in rover localization. The effectiveness of the proposed method is demonstrated through numerical simulations using the map of planned landing site.

\subsection{Related Work}
\subsubsection{Multi-agent systems in terrestrial applications}
There are a lot of studies on multi-vehicles for both ground and aerial systems. Compared to ground vehicles, typically, aerial vehicles can cover larger search areas in a given time. For UAV teams, \cite{c_path_1} studies use of Dijkstra algorithm for real-time motion control. \cite{c_path_2} utilizes multiple active sensors to avoid mid-air collisions. 
Recent methods study decentralized visual simultaneous localization and mapping (SLAM) (e.g., \cite{c_path_3}) and planning under uncertainty (e.g., \cite{posmpd}). 
However, UAVs are typically highly resource limited, ranging from their sensory limitation in their localization accuracy of ground targets as well as computational and battery charge limitations.
Multi-ground/aerial system is focused on a combined system to address some of these limitations \cite{rc1,rc2}.

\subsubsection{Multi-agent systems in planetary exploration}
Unlike terrestrial exploration, Mars surface makes it more difficult for the exploration vehicles to estimate their locations.
\cite{PlanetMars1} describes an integrated system for coordinating multi-rover behavior with the overall goal of collecting planetary surface data. \cite{PlanetMars2} shows how to construct evaluation functions in dynamics, noisy and communication-limited collective environments.
\cite{Nilsson2018} focuses on navigation and coordination of the Mars rover and copter teams such as the one shown in Fig.~\ref{fig:8_rc}, and aims at bridging the gap between high-level mission specifications and low-level navigation and control techniques under the environment uncertainty.
Also, there is a large body of work studying the Mars rover/orbiter collaboration and data sharing (e.g., \cite{maxRAL})

\subsection{Mission Concept}
Our conceptual mission treats the copter as a scout to collect data before the beginning of rover's traverse on each Martian day. The rover plans an optimal path that avoids critical terrain hazards and regions with low localizability using the updated map information by the copter. We assume a copter deployment scenario similar to the upcoming Mars 2020 rover/copter mission. The copter is first deployed from the rover to the ground, and then it takes off directly from the ground after the rover drives away for a certain distance. This is to protect the precious rover from accidental damage caused by copter's misbehavior. Mission assumptions are:
\begin{itemize}
\item The copter can only fly 90 seconds per day due to power constraints \cite{Balaram2018}
\item The rover has more energy and operation time but can only traverse at a very low speed ($\sim$4 cm/s)
\item The copter can move fast and is not hindered by obstacles
\item The copter cannot get closer than a certain distance to the rover for safety.
\end{itemize}

\subsection{Contributions}
In this work, we aim to develop a joint rover and copter motion planner that minimizes the localization uncertainty of the rover in a noisy localizability map. The contribution of this paper is summarized as follows:
\if0
\kyon{Will come back to this later....}
\begin{itemize}
    \item Formulate a perception-aware navigation problem for multi-asset Mars missions
    \item Propose an active mapping solution for a secondary robot (i.e., copter) to support the main mission of a primary robot (i.e., robot)
    \item Perception-aware planning with application to Mars rover-copter navigation
    \item Active mapping 
    \item Implement an efficient online solver that improves the probabilistic worst case
\end{itemize}
\fi
\begin{itemize}
    \item Define hyper-belief map representation for the VO odometry error uncertainty;
    \item Solve the copter's where-to-map problem for minimizing the rover localization uncertainty% in the odometry error map;
    \item Solve for optimal image capture point for the copter, while considering the trade-off between field-of-view and resolution of the camera;
    \item Propose an iterative rover/copter path planning method and demonstrate the effectiveness through simulations.
\end{itemize}

\section{Problem Description}
\begin{figure}[b]
	\centering
	\includegraphics[width=1.0\columnwidth]{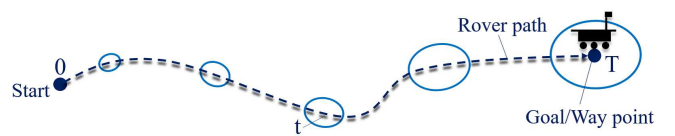}
	\vspace{-7mm}
	\caption{Belief propagation in rover's state estimation. Since odometry error is dependent on terrain localizability, the belief evolution depends on the rover's path.}
	\label{fig:problemstatement}
\end{figure}

The main problem that we are going to address is a two-agent motion planning problem where one agent is expected to assist the other for the mission to succeed. The objective of the primary agent is to reach the goal with high localization confidence, while the secondary agent works as a scout to increase the information quantity about the environment to help primary agent's motion planning. In our problem setup, two agents are Mars rover and Mars helicopter. The prior imperfect information comes from the Mars orbiter.

\pr{Primary agent (rover)}
Rovers' mobility is limited to relatively flat surfaces. We denote the rover state by $ x^{r} $ which is typically parameterized by a position in X-Y plane and its heading. The rover moves with a slow speed traveling from the initial state $ x_{0}^{r} $ to the goal state $ x_g^{r} $.
Due to the absence of global positioning methods on Mars, a rover has to estimate its pose incrementally by propagating relative transformations between infrequent pose updates available through local bundle adjustment or satellite image matching.  This incremental dead-reckoning approach is prone to error accumulation. Typically, such error is captured by propagating the covariance of state estimation as shown in Fig.~\ref{fig:problemstatement}. The rover has to keep its state uncertainty small enough to avoid obstacle collisions or entering keep-out-zones.

\pr{Secondary agent (copter)}
The Mars copter has an ability to fly and explore larger areas with down-facing cameras. The copter state is denoted by $ x^{c} $ which is composed of a 3D position and a heading angle. The difference in altitude affects the quality of sensor measurements such as resolution and estimation confidence, as well as the size of the area it can observe.
While copter is much faster than the rover, its  operation time is limited to a single 90-second flight per day due to the severe power constraints. In addition, the copter's should constantly keep a certain distance from the rover to satisfy the safety constraint for the rover.

\pr{Environment}
%The environment is not fully known a priori in case of planetary exploration. 
On parts of Mars, we have surface data available from satellite imagery up to 25cm/px resolution. This limited information needs to be complemented by onboard sensing for any surface navigation.
For the current navigation methods, the terrain type is an important factor to determine localizability. In case of \ac{WO}, significant error is expected on loose sandy terrain due to wheel slippage, and \ac{VO} is affected by visual texture on the ground. Therefore, a localizability index can be computed using terrain classification techniques such as \cite{Rothrock2016}. However, the  resolution of satellite data is not high enough to confidently assess the localizability on the Mars surface. Adding copter's data helps increasing the accuracy of localizability information with increased resolution and contextual information.

\begin{figure}[t]
    \centering
	\setlength{\subfigtopskip}{-2pt}
	\setlength{\subfigcapskip}{-3pt}
	\setlength{\abovecaptionskip}{-3pt}
    \subfigure[The propagation of state hyper-belief along a path. The yellow region represents the size of the state covariance in a distribution of error ellipse. This is described as a donut-shaped region around error ellipse.
    ]{\includegraphics[width=1.0\columnwidth]{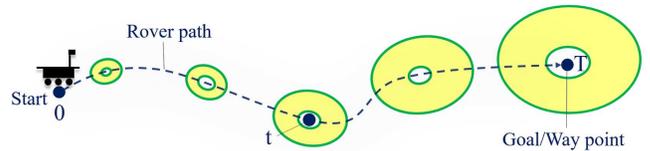}}\\
    \subfigure[Hyper-belief update after copter's measurement. The covariance of terminal hyper-belief is reduced due to the localizability update on the rover path.]{\includegraphics[width=1.0\columnwidth]{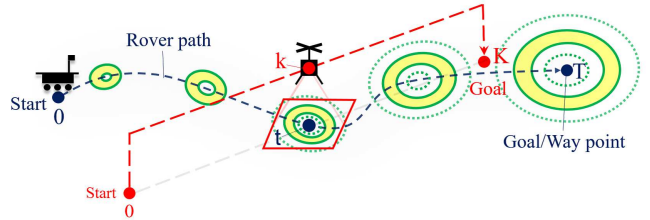}}
	\caption{Hyper-belief propagation in rover's state estimation.}
	%\vspace{-3mm}
	\label{fig:problemstatement2}
\end{figure}

\pr{Problem}
Given the partial prior map from satellite, we simultaneously solve for the rover path planning and optimal locations for copter measurements acquisition to minimize the state uncertainty accumulation in rover localization (i.e., error ellipse in Fig.~\ref{fig:problemstatement}). The main challenge here is that we cannot precisely predict how the covariance evolves along a given path \textit{before} traversal. What we can do instead is to have a belief on covariance evolution, given the prior localizability estimate. We call this a \textit{hyper-belief} as it is a belief over state beliefs.
The hyper-belief propagation process is depicted in Fig.~\ref{fig:problemstatement2}. In the top figure, the hyper covariance along a path is shown by yellow regions, which is regarded as the distribution over the state error ellipse. To reduce the size of the terminal covariance, we specifically consider two approaches: 1) plan a rover's path to avoid excessive error accumulation, and 2) update the localizability map by copter measurements to change the hyper-belief evolution along paths. The second approach is presented in Fig.~\ref{fig:problemstatement2}b. After the map update by the copter, the uncertainty in covariance evolution is reduced. As a result, the covariance in hyper-beliefs becomes smaller (the area of yellow region), and hence the confidence on state uncertainty increases.

\section{Localization Error Propagation}
In this section, we describe how we model the propagation of pose estimates and their uncertainty, along with how we encode localizability into a map.

\subsection{Pose belief propagation}
To capture uncertainty in the robot pose, we directly work and plan over pose beliefs. A pose belief is a probability distribution over the possible states of the robot. Below, we will present the mathematical definition of pose beliefs and propagation process. 

\pr{Stochastic pose representation}
We treat the pose of the rover as a member of Lie group $ T \in SE(3) $, which encodes the rover's 3D pose and orientation in a $4 \times 4$ transformation matrix. To capture the stochasticity of the rover pose in $SE(3)$, we rely on perturbation models on this Lie group \cite{Barfoot2014}. Denote the small pose perturbation by $\xi=(\rho, \phi)\in\mathbb{R}^6$, where $\rho\in\mathbb{R}^3$ is perturbation in position and $\phi\in\mathbb{R}^3$ in orientation. Using this perturbation representation, the noisy rover pose is expressed as 
\begin{align}\label{eq:stochasticPose}
    T \equiv \exp(\xi^{\wedge}) \bar{T},
\end{align}
where $ \bar{T} \in SE(3) $ is the nominal noise-free pose, and the noise vector $\xi$ is mapped to a member of $SE(3)$ using the exponential map through the Lie algebra.
Note that $\exp(\cdot)$ is an operator producing the matrix exponential which gives the connection between a matrix Lie algebra and the corresponding Lie group.
More precisely, we map a vector $\xi$ to a member of the Lie algebra using the operator $\wedge$
\begin{align}
    \xi^{\wedge}
    = \begin{bmatrix} 
    \phi^{\times} & \rho \\
    0^{T} & 0
    \end{bmatrix}
\end{align}
where the operator $\times$ produces a skew-symmetric matrix from a vector, i.e.,
\begin{align}
    \bm{\phi}^{\times} = \begin{bmatrix}
    \phi_1 \\ \phi_2 \\ \phi_3
    \end{bmatrix} ^{\times}
    = \begin{bmatrix}
    0 & -\phi_3 & \phi_2 \\
    \phi_3 & 0 & -\phi_1 \\
    -\phi_2 & \phi_1 & 0
    \end{bmatrix}
    \:.
\end{align}
This perturbation approach ensures that noisy variables also belong to the same group (i.e., $ T, \bar{T} \in SE(3) $). Equation \eqref{eq:stochasticPose} can be interpreted as a noise injection into the group $ SE(3) $. 
We assume that the perturbation follows a zero-mean Gaussian distribution
\begin{align}\label{eq:Gaussian_perturb}
    \xi \sim \mathcal{N}(0, \Sigma)
\end{align}
with covariance matrix $ \Sigma \in \mathbb{R}^{6 \times 6} $. 
In the sequel, we use the pair $\{\bar{T}, \Sigma \} $ to characterize the distribution of the stochastic pose.
\pr{Pose belief evolution.}
Using the above definition, we represent the robot pose and covariance at time $ t $ by $ \{ \bar{C}_{t}, P_{t} \} $, and the relative transformation and its uncertainty between time $ t $ and $ t+1 $ by $ \{ \bar{T}_{t+1, t}, \Sigma_{t+1, t} \} $. Similar to the work \cite{Barfoot2014}, we propagate the mean and covariance of the robot pose up to the second-order term:
\begin{align}
    \bar{C}_{t+1} &= \bar{C}_{t} \bar{T}_{t+1, t} \label{eq:pose_propagation} \\
    P_{t+1} &= P_{t} + \mathcal{T}_t \Sigma_{t+1, t} \mathcal{T}_t^{T} \label{eq:uncertainty_propagation}
\end{align}
where the adjoint $ \mathcal{T}_t = \mathrm{adj}(\bar{C}_{t}) \in\mathbb{R}^{6 \times 6}$ can be computed from the current pose. Note that $\mathrm{adj}(\cdot)$ is an operator producing the adjoint of an element from the Lie group for rotations and poses.
\subsection{Map representation}
We approximate the environment as a map represented by a regular grid with $n$ cells. This allows us to analyze the localizability, or equivalently the expected perturbation, for the individual cells.
Localization methods such as \ac{WO} and \ac{VO} are always contaminated with odometry error or perturbation $\xi$ as in Eq.~\eqref{eq:stochasticPose}. It is reasonable to assume that the intensity of odometry error, i.e., $ \Sigma=Cov[\xi] $, depends on environmental factors such as terrain types. As our focus is on the path selection in a global scale, we simplified the error model by assuming the error accumulation is independent of local motion types and is proportional to the distance traveled. With this, we model the environment as a set of normalized motion estimation covariances, denoted by $ m = \{ \tilde{\Sigma}_1, \cdots, \tilde{\Sigma}_n \} $. Simply put, if the rover traverses on $i$-th cell for $d$ meters, the covariance of motion estimation becomes $d\tilde{\Sigma}_i$ and state estimation uncertainty grows accordingly with Eq.~\eqref{eq:uncertainty_propagation}. Thus, we approximate the environment with a regular grid composed of $n$ cells, and associate with each cell  a localizability index.

The localizability map cannot be built before actual traversal on every location. Instead, we keep a probability distribution (belief) on map $m$ based on remote sensing data. In our problem, we build a map belief based on satellite measurements $z^{s}$, copter measurements $z^{c}_{k}$ and position $x^{c}_{k}$:
\begin{align}
    b^{m}_{k} = p(m | z^{s}, z^{c}_{1:k}, x^{c}_{1:k})
\end{align}
where $z^c_{1:k} = \{z^c_1, \cdots, z^c_k\}$ and $x^c_{1:k} = \{x^c_1, \cdots, x^c_k\}$. 
By choosing to model the environment as a collection of cells, the belief  $b^{m}_{k}$ contains $n$ individual beliefs. 
\begin{align}
    b^{m}_{k} &\approxeq \{ b^{m^{1}}_{k}, \cdots, b^{m^{n}}_{k} \} \\
    b^{m^{i}}_{k} &= p(m^{i} | z^{s}, z^{c}_{1:k}, x^{c}_{1:k})
\end{align}
Specifically, the prior map belief obtained from satellite data is represented as $b^{m^{i}}_{0} = p(m^{i} | z^{s})$.
Note that modelling the belief as a collection of beliefs over map cells offers a computationally-tractable approximation to the evolution of the full joint map belief.
\subsection{Pose hyper-belief propagation}
The pose belief is propagated with Eqs.~\eqref{eq:pose_propagation}~and~\eqref{eq:uncertainty_propagation} if the relative transformation is given in a deterministic form. In our problem, we plan rover's path on a map belief and the uncertainty accumulation only given as a stochastic form. Given an initial pose covariance $P_0$, and the rover's path $X^{r}$ expressed as a list of cell indices, the pose covariance $P_{g}$ at the terminal pose $x^{r}_{g}$ is predicted as
\begin{align}
    P_{g} &= P_{0} + \sum_{i \in X^{r}} d_{i} \mathcal{T}_{i}  \tilde{\Sigma}_{i} \mathcal{T}_i^{\top}
    \label{eq:hyper_belief_propagation}
\end{align}
where $d_i$ is the travel distance on $i$-th cell. 
At time instant $k$, the local motion estimation variance $\tilde{\Sigma}_{i}$ is unknown and modeled with the belief distribution $b^{m^i}_k$.
%%%%%%%%%%%%%%%%%%%%%%%%%%%%%%%%%%%%%%%%%%%%%%%%%%%%%%%%%%%%%%%%%%%%
\vspace{-1mm}
\section{Map Update}
The localizability map is first initialized with the satellite measurements, and then dynamically updated by the copter. In this section, we describe the measurement models and the prediction of map belief update by future copter measurements.
\subsection{Measurements}
\begin{figure}[t]
	\centering
	\includegraphics[width=0.8\columnwidth]{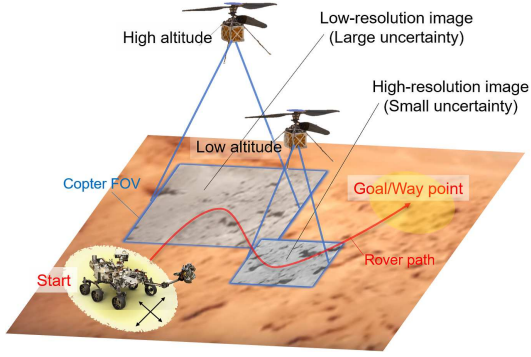}
	\caption{Trade-off between FOV and resolution of copter measurements.}
	\vspace{-3mm}
	\label{fig:FOV}
\end{figure}

In Mars missions, there are various sensing modalities, including measurements from orbiter, copter, and rover's onboard sensors. Each sensor differently contributes to the map update with various resolution and confidence.
\pr{Satellite observation}
The satellite observation is considered to be the lowest in resolution and fidelity, but the highest in areal coverage. We assume that all locations in the map are measured by satellites. The satellite-based measurements usually come with large uncertainty. The uncertainty is also terrain type dependent. %: it is relatively easy to identify large boulder fields, but some terrains could be confusing to each other. 
\pr{Copter observation}
The copter observation has higher resolution and fidelity than satellites, but the scope is smaller than the entire map. The intensity of measurement noise varies with the copter's 3D location as illustrated in Fig.~\ref{fig:FOV}. The blue square in this figure represents the field of view (FOV) of the copter's camera.
At higher altitudes, copter can observe larger regions but the noise intensity is higher. In contrast, at lower altitudes, the copter can observe smaller regions but more accurately. In addition to the copter altitude, its 2D location can cause variation in noise intensity depending on the texture of the underlying terrain. 
\pr{Copter Measurement Model}
The copter's measurement model for the $i$-th map cell is described as a probabilistic distribution over copter measurements $z^{c}$ given copter pose $x^{c}$ and map cell $m^{i}$
\begin{align}
    p(z^{c} | m^{i}, x^{c}) .
\end{align}
For simplicity, we assume that the copter can directly observe the localizability index of a location, and the Gaussian noise is injected to the measurement:
\begin{align}
    p(z^{c} | m^{i}, x^{c}) = \mathcal{N}(m^{i}, w(x^c))
    \label{coptermodel1}
\end{align}
where $w(\cdot)$ is a function representing the noise intensity that varies depending on the 3D location of the copter such as altitude.
\pr{Map update}
After each copter measurement, we update the map $ m $; precisely speaking, we update the beliefs over map cells using the Bayesian update rule:
\begin{align}
    b^{m^i}_{k+1} = \tau( b^{m^i}_{k}, z^{c}_{k+1}, x^{c}_{k+1} )  .
    \label{mapupdate}
\end{align}
We discuss this update rule further in the next subsection. 

\subsection{Predictive map update}\label{sec:map_update}
To understand how the map belief will evolve, it is essential to evaluate and predict the impact of  the copter observations on the map. 
\pr{Belief over copter measurement}
Given the prior knowledge, we can model the distribution of future copter observations on the $i$-th cell by marginalizing over the prior map distribution:
\begin{align}
    b^{z^{c}}_{k+1}
    &= p({z}^{c}_{k+1} | z^s, z^{c}_{1:k}, x^{c}_{1:k+1}) \\
    &= \int p(m^i | z^{s}, z^{c}_{1:k}, x^{c}_{1:k}) p(z^{c}_{k+1} | m^i, x^{c}_{k+1}) dm^i  .
    \label{eq:zckzsxc}
\end{align}

\pr{Predictive map update}
Based on the belief over copter's measurement, we predict the map belief after the measurement by:
\begin{align}
b_{k+1}^{m^i} = \tau(b^{m^i}_k,z^{c,ml}_{k+1},x^c_{k+1}),~~~z^{c,ml}_{k+1}=\arg\max b^{z^c}_{k+1}
\end{align}
Note that we use the most likely copter measurement $z^{c,ml}_{k+1}$ to predict the map belief. This recursion can be computed as:
\begin{align}
    b^{m^i}_{k+1} &= 
    p(m^i | z^s, z_{0:k+1}^c , x_{0:k+1}^c)\\
    &= \frac{p(z_{k+1}^c | m^i, z^s, z_{0:k}^c , x_{0:k+1}^c)p(m^i | z^s, z_{0:k}^c , x_{0:k}^c)}{p(z_{k+1}^c | z^s, z_{0:k}^c , x_{0:k+1}^c)}\\
    &= \eta p(z_{k+1}^c | m^i , x_{k+1}^c)b^{m^i}_{k}
\end{align}
We assume Gaussian distributions provides a good approximation for the map belief. This assumption can lead to a computationally tractable update rule, given $b_{0}^{m^i} = p(m^i | z^s) = \mathcal{N}(\mu^i, \sigma^{i^2})$ and Eq.~\eqref{coptermodel1}. 

%Based on the belief over copter's measurement, we predict the map belief after the %measurement
%\begin{align}
 %   \grave{b}^{m^i}_{k+1} = \tau'( b^{m^i}_{k}, b^{z^{c}}_{k+1}, x^{c}_{k+1} ) .
%\end{align}
%Note that we allow only one update prediction for each cell, and we typically take the highest %confidence measurement if there are multiple measurements planned for the same location. %Let's say we have multiple measurements planned for the $i$-th map cell. We only take %highest confidence measurement to predict the updated map belief.
%\pr{Gaussian assumption}
%For ease of computation, we assume the Gaussian distribution for the prior map beliefs
%\begin{align}
  %  b^{m^i}_{0} = p(z^s, x^{c}) = \mathcal{N}(\mu_i, \sigma_i^2).
 %   \label{priormodel}
%\end{align}
%Based on the prior distribution, the measurement belief is obtained as
%\begin{align}
  %  b^{z^{c}}_{1} =
 %   p(z^{c} | z^s, x^{c}) = \mathcal{N}(\mu_i, \sigma_i^2 + w(x^c))
%\end{align}
%using the sensor model in Eq.~\eqref{coptermodel1}. 
%Using the Bayesian update, the updated map is predicted as
%\begin{align}
%\grave{b}^{m^i}_{1}
%=p(z^{c} | z^s, x^{c}) p(z^s, x^{c}) 
%=\mathcal{N}\left( \mu_i, \cfrac{\sigma_i^2 (\sigma_i^2 + w(x^c))}{2\sigma_i^2 + w(x^c)}\right).
%\end{align}

%%%%%%%%%%%%%%%%%%%%%%%%%%%%%%%%%%%%%%%%%%%%%%%%%%%%%%%%%%%%%%%%%%%%
\section{Planning}
In this section, we present an approximate joint-space search method for rover's path and copter's measurements given a prior localizability map from the satellite. The ultimate objective is to navigate the rover to the goal with low localization error accumulation. 
\begin{algorithm}[!t]
\caption{Iterative Rover/Copter Optimization}
\label{alg:2}
Generate initial map belief from satellite measurement $z^{s}$\\
{\bf For}~$j=1$ {\bf to} max iteration
\begin{description}
\item Find the optimal rover path $X^{r*}_{j}$ given map belief (Section~\ref{sec:r})
\item Find the optimal copter path $X^{c*}_{j}$ given map belief and rover path (Section~\ref{sec:c})
\item Update map belief based on copter measurements (Section~\ref{sec:map_update})
\end{description}
{\bf Return}~rover and copter paths $({X}^r, X^{c})$
\end{algorithm}
\subsection{Approximate Joint-space Search}
Searching the joint space of rover path and copter measurements is computationally demanding. We introduce an approximate solution to the joint-space search.
\pr{Search space}
Let $X^{r}=\{ x^r_{0}, x^r_{1}, \cdots, x^r_{g} \}$ denote a smooth rover path that connects start and goal positions, and $X^{c}=\{ x^c_{1}, x^c_{2}, \cdots \}$ denote ordered copter measurement locations. The entire search spaces are represented by $\mathcal{X}^{r}$ and $\mathcal{X}^{c}$, respectively. The joint space is defined as a product of two spaces, $\mathcal{X}^{r} \times \mathcal{X}^{c}$. 
\pr{Optimization problem}
The joint-space optimization problem is formulated as follows. Given a cost function $J$ and prior map belief $b^{m}_{0}$, find optimal $X^{r}$ and $X^{c}$ that minimize $J$:
\begin{align}
    (X^{r*}, X^{c*}) = 
    \argmin_{(X^{r}, X^{c}) \in \mathcal{X}^{r}\times\mathcal{X}^{c}} J(X^{r}, X^{c}; b^{m}_{0})
\end{align}
The detail of the cost function is given in the next section.
\pr{Iterative programming}
Searching the full joint space is computationally demanding. We introduce an approximate solution to the joint-space search by decomposing the problem into two: rover path planning and copter perception planning. We approximately solve the optimization problem by iteratively solving these two problems until it converges (see Algorithm~\ref{alg:2}).

\subsection{Optimization Criteria}
In this section, we discuss the optimization objective mathematically. 

\pr{Cost for pose belief}
We first define a scalar cost $\Omega$ for evaluating the terminal pose belief after driving path $X^{r}$. Consider the pose belief represented by a pair $\{\bar{C}_g, P_g\}$. We model the cost for the pose estimate by computing the trace of covariance matrix:
\begin{align}
    \Omega = \tr(P_g)
\end{align}
Note that our objective is to reduce the accumulation error in rover pose estimation, i.e., minimizing $\Omega$.

\pr{Cost for pose hyper-belief}
Due to the uncertainty in the localizability map, the pose covariance cannot be obtained in a deterministic form prior to driving. Instead, it is given as a stochastic prediction as formulated in Eq.~\eqref{eq:hyper_belief_propagation}. Accordingly, the cost $\Omega$ is given as a random variable which is computed from a map belief as
\begin{align}
   \Omega = \tr(P_0) + \sum_{i \in X^{r}} d_{i} \tr\left( \mathcal{T}_{i}  \tilde{\Sigma}_{i} \mathcal{T}_i^{\top} \right)
    \label{eq:criteria}
\end{align}

To determinize the cost, we introduce the inverse cumulative distribution function $\Phi_\Omega^{-1}(p)$ for $\Omega$ that returns a threshold value at which the cumulative probability of random variable is equal to the given probability $p$. We define the cost function as
\begin{align}
    J(X^{r}, X^{c}; b^{m}_0, p) = \Phi_\Omega^{-1}(p) .
    \label{eq:objective}
\end{align}
The threshold parameter $p$ can be arbitrarily set. Qualitatively, by selecting a higher value, the planner tries to minimize the worst-case cost, while a smaller value (e.g., 0.5) optimizes for nominal performance.

\subsection{Rover Planning}\label{sec:r}

The objective of rover path planning is to solve the optimal path that minimize the objective function in Eq.~\eqref{eq:objective} given a map belief. As presented in Algorithm~\ref{alg:2}, the rover path planning is based on the current best estimate of map belief that takes future copter measurements into account. Let $X^{c}_{j}$ be the optimal copter measurements after $j$-th iteration. The $(j+1)$-th rover pose optimization is formulated as
\begin{align}
\label{optimal_rover_path}
    X^{r*}_{j+1} = 
    \argmin_{X^{r} \in \mathcal{X}^{r}} J(X^{r}; X^{c}_{j}, b^{m}_{0})
\end{align}
or more directly,
\begin{align}
    X^{r*}_{j+1} = 
    \argmin_{X^{r} \in \mathcal{X}^{r}} J(X^{r}; \grave{b}^{m}_{K})
\end{align}
where the predictive map belief $\grave{b}^{m}_{K}$ is built based on the prior and copter's $K$ measurements.
We solve this problem by casting it as a graph search problem. Objective criteria in Eq.~\eqref{eq:criteria} allow us to set positive uncertainty accumulation costs to graph edges. The minimum uncertainty accumulation path can be computed by the Dijkstra's shortest path algorithm.

\subsection{Copter Planning}\label{sec:c}

The copter's perception planning problem also aims at improving the objective function in Eq.~\eqref{eq:objective}. Although the rover localization error has no direct connection to the copter's measurements, the updated localizability map by copter's measurements can improve localization error by indirectly affecting rover's path planning process.

\pr{Objective}
We rewrite the objective function as the difference in cost by incorporating copter's measurements $X^{c}$
\begin{align}
\label{optimal_copter_path}
    X^{c*}_{j+1} = 
    \argmin_{X^{c} \in \mathcal{X}^{c}} J'(X^{c}; X^{r}_{j+1}, b^{m}_{0})
\end{align}
where
\begin{align}
    J'(X^{c}; X^{r}, b^{m}_{0}) 
    &= J(X^{c}; X^{r}, b^{m}_{0}) - J(\O; X^{r}, b^{m}_{0}) 
    \label{eq:copter_objective}  \\
    &= -\sum_{i \in X^{r}} \max_{k} v_{ik}.
\end{align}
Note that $v_{ik}$ is the cost improvement in the $i$-th cell along rover's path by copter's $k$-th measurement. If the $i$-th location is not observed by any copter measurement, the cost improvement is zero. If the location is observed by multiple measurements, the most effective update is taken. 

\pr{Constraints}
There are severe operational constraints for the Mars helicopter. In this paper, the following constraints are considered.
\begin{align}
  &h_{\min} \leq x_{z}^c \leq h_{\max}\label{eq:stc1}\\
  &0 \leq K \leq k_{\max}\label{eq:stc2}\\
  & d(x^c_K, x^{r}_{0:T}) > \Delta
  \label{eq:stc3}
\end{align}
where $x_{z}^c$ is the altitude of the copter.
The first constraint in Eq.~\eqref{eq:stc1} comes from both copter FOV constraint and camera resolution constraint.
This constraint has the trade-off between FOV and resolution of the camera.
In Eq.~\eqref{eq:stc2}, the limitation of the copter battery constrains copter's flight time or distance. The positive scalar $k_{\max}$ is the maximum time step of the copter and should be larger than the final step of the copter $K$.
The final constraint in the planning framework comes from the copter position at the $K$ (final step) in Eq.~\eqref{eq:stc3}. Parameter $\Delta$ is the distance margin between rover and copter for their safety.

\pr{Search Method}
To efficiently solve the minimization problem in Eq.~\eqref{eq:copter_objective}, we formulate the problem as a variant of knapsack problem. We discretize the copter's state space as $\mathcal{X}^{c}_{search}$. Then, we find the optimal combination of measurements that maximize the total value while satisfing various constraints. The value of a measurement is computed as the sum of cost improvement along a path
\begin{align}
    V_k = \sum_{i \in X^{r}} v_{ik} .
    \label{eq:value}
\end{align}
Analogous to the capacity constraint of knapsack problem, this problem poses the flight distance constraint 
\begin{align}
    \mathrm{PathLength}(X^{c}) \leq l_{\max} .
\end{align}
Note that, unlike the capacity constraint, the distance constraint is order-sensitive. Therefore, before testing the constraint, we solve the travel salesman problem (TSP) to sort the locations so that the entire path has shortest distance.
The algorithm for the TSP-based knapsack problem is presented in Algorithm \ref{alg:1}.

\begin{algorithm}[!t]
\caption{TSP-based knapsack problem}
\label{alg:1}
%\DontPrintSemicolon
%\BlankLine
Maximum path length $l_{\max}$\\
Set initial cost $c_{\max}=0$\\
{\bf For}~$j=1$ {\bf to} max iteration
\begin{description}
\item $X^{c} = \mathrm{SamplePathWithTSP}()$
\item $c = Cost(X^{c})$
\item $l = PathLength(X^{c})$
\item {\bf If}~{$c>c_{\max}$} and {$l \leq l_{\max}$}
\begin{description}
\item Update max cost $c_{\max} = c$
\item Store optimal copter path $X^{c*}=X^{c}$
\end{description}
\end{description}
{\bf Return} Optimal copter path $X^{c*}$\;
\end{algorithm}

%%%%%%%%%%%%%%%%%%%%%%%%%%%%%%%%%%%%%%%%%%%%%%%%%%%%%%%%%%%%%%%%%%%%
\section{Simulation Results}
\begin{figure}[!t]    
    \centering
	\includegraphics[width=1.0\columnwidth]{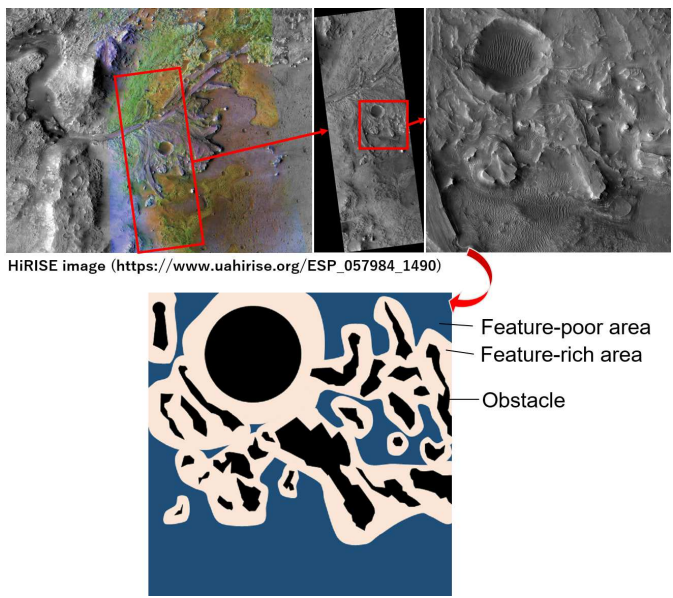}
	\vspace{-2mm}
    \caption{Mars map by the HiRISE.}
    \vspace{-5mm}
	\label{fig:8_map1}
\end{figure}
In our simulation, we use the HiRISE map of Jezero crater (see Fig.~\ref{fig:8_map1}), the landing site for the Mars 2020 mission. We manually annotated the map for feature-rich and poor areas in the localizability context.

\subsection{Single-run Simulation Results}
Under the deployment scenario discussed earlier, the start and goal positions of two agents are different. Assuming the bottom-left corner of the map in Fig.~\ref{fig:8_map1} represents $ (0,0,0) $ coordinate, the initial positions of the rover and copter are set to $(20,~20)$ and $(30,~70,~2)$, and the goal positions of the rover and copter are set to $(140,~140)$ and $(120,~140,~2)$, respectively.

\begin{figure}[!t]    
    \centering
	\setlength{\abovecaptionskip}{-3pt}
	\includegraphics[width=1.0\columnwidth]{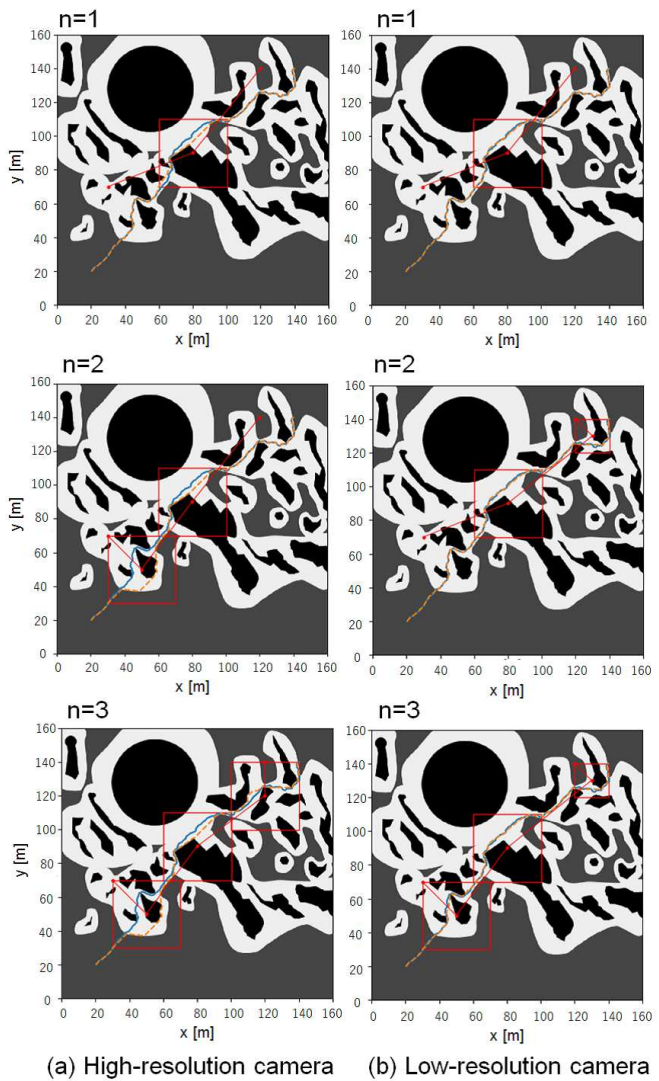}
	%\vspace{-8mm}
    \caption{Where-to-map and iterative rover path results updated by (a) the high-resolution camera and (b) the low-resolution camera of the copter.}
    \vspace{-2mm}
	\label{fig:8_highlow}
\end{figure}

Figure~\ref{fig:8_highlow} shows the selected rover paths and copter measurements planned by the proposed method under different assumptions on measurement opportunities and qualities. The rover and copter are operated on 800-by-800 grid in 0.2\,[m] resolution. $n$ denotes the number of the images copter can take. The resolution of the camera influences the amount of uncertainty reduction given the measurement altitude. The blue solid line and the orange dashed line represent the initial rover path and the updated rover path using iterative planner, respectively. The red solid line shows the copter body path. The red point and the red square represent the copter position and FOV of the copter's camera, respectively.

The observation positions (OP) $\#1$, $\#2$, and $\#3$ in Table~\ref{t1} denote the locations where the copter will take images. Table~\ref{t1} also shows the associated reduction rates of the rover's pose hyper-belief in Eq.~\eqref{eq:criteria}.
Note that we show two types of copter's updates with high- and low-resolution cameras.
The reported results show the reduction in rover's worst-case odometry error uncertainty. As the number of images captured by the copter increases, the rover uncertainty decreases.

\begin{table}[!t]
\caption{Observation positions (OP) by the copter and reduction rates of rover's pose hyper-belief}
\vspace{-3mm}
\centering
\begin{tabular}{c @{\quad} c @{\quad} c @{\quad} c @{\quad} c}
\hline
$n$ & OP$\#1$ [m] & OP$\#2$ [m] & OP$\#3$ [m] & Reduction rate [$\%$]\\
\hline
\multicolumn{4}{l}{(High-resolution updates)}\\
$1$ & (80,~90,~10) & - & - & 14.15\\
$2$ & (50,~50,~10) & (80,~90,~10) & - & 27.49\\
$3$ & (50,~50,~10) & (80,~90,~10) & (120,~120,~10) & 40.66\\
\hline
\multicolumn{4}{l}{(Low-resolution updates)}\\
$1$ & (80,~90,~10) & - & - & 7.35\\
$2$ & (80,~90,~10) & (130,~130,~5) & - & 14.19\\
$3$ & (50,~50,~10) & (80,~90,~10) & (130,~130,~5) & 20.93\\
\hline
\end{tabular}
\vspace{-3mm}
\label{t1}
\end{table}
\subsection{Monte Carlo Simulation Result}
\begin{figure}[!t]    
    \centering
	\includegraphics[width=1.0\columnwidth]{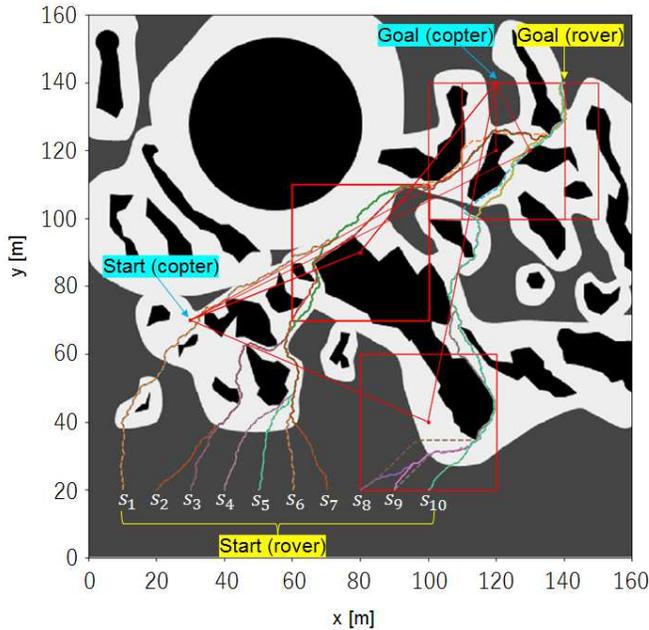}
	\vspace{-2mm}	
    \caption{Monte Carlo Simulation result with various initial rover's positions.}
    \vspace{-3mm}
	\label{fig:mcresult}
\end{figure}

This subsection shows a Monte Carlo simulation result of collaborative rover and copter planning. The number of copter measurements is fixed to $n=1$. 
We ran a simulation for each initial position of the rover $s_1, s_2,\cdots, s_{10}$ listed in Table~\ref{t2}. We also compared our algorithm with the baseline random mapping approach in which the copter randomly selects a point on the rover's trajectory. We present the average result of 100 runs for the baseline. Figure~\ref{fig:mcresult} and Table~\ref{t2} show the optimal paths of rover and copter and reduction rates by copter observations. Nominally, we obtained 10--20\% gain in uncertainty reduction with the proposed algorithm, while the random mapping approach only gives less than 10\% gain, highlighting the effectiveness of the proposed approach in reducing rover's localization error by utilizing the limited resource of the copter.

\begin{table}
\caption{Reduction rates in MC simulation result}
\vspace{-3mm}
\centering
\begin{tabular}{c @{\quad} c @{\quad} c @{\quad} c}
\hline
\multicolumn{2}{c}{Rover's initial position [m]} &
\multicolumn{2}{c}{Reduction rate [$\%$]}
\\
\multicolumn{2}{c}{} &
Proposed &
Baseline
\\
\hline
$s_1$ & (10,~20) & 13.36 & (6.04)\\
$s_2$ & (20,~20) & 14.15 & (6.59)\\
$s_3$ & (30,~20) & 15.12 & (7.25)\\
$s_4$ & (40,~20) & 15.83 & (7.73)\\
$s_5$ & (50,~20) & 16.37 & (8.16)\\
$s_6$ & (60,~20) & 16.42 & (7.62)\\
$s_7$ & (70,~20) & 15.67 & (7.36)\\
$s_8$ & (80,~20) & 18.38 & (8.39)\\
$s_9$ & (90,~20) & 18.18 & (9.51)\\
$s_{10}$ & (100,~20) & 18.41 & (10.19)\\
\hline
\end{tabular}
\label{t2}
\end{table}

%%%%%%%%%%%%%%%%%%%%%%%%%%%%%%%%%%%%%%%%%%%%%%%%%%%%%%%%%%%%%%%%%%%%
\section{Conclusion}
In this paper, we addressed the copter's where-to-map problem for reducing the rover odometry error uncertainty using the introduced hyper-covariance concept. We proposed a copter path planner for determining the points where the copter will take images while considering the trade-off between the field-of-view and resolution of its camera. An iterative rover/copter path planning method is proposed to solve the problem. Finally, we have demonstrated the effectiveness of the proposed method through the numerical simulation results.
The battery of Mars copter is strictly limited since the flight on Mars needs high-speed rotor which will consume a lot of power. 
This leads to strict time constraints on \textit{when} the copter flies. As a future work, we will consider {\it when-to-map} problem by the copter and conduct an experiment on the collaborative rover/copter system.

%\section*{Acknowledgment}

%\ifCLASSOPTIONcaptionsoff
 % \newpage
%\fi

%\bibliographystyle{IEEEtran.bst}
%\bibliography{refs}

\end{document}